\documentclass[runningheads]{llncs}
\usepackage{graphicx}
\usepackage{times}
\usepackage{epsfig}
\usepackage{graphicx}
\usepackage{amsmath}
\usepackage{amssymb}
\usepackage{booktabs}
\usepackage{tablefootnote}
\usepackage{enumitem}
\usepackage{multicol}
\usepackage{subfigure}
\usepackage{tabularx}
\usepackage{bbding}
\newcommand{\myparagraph}[1]{\vspace{-8pt}\paragraph{#1}}

\newcolumntype{L}[1]{>{\raggedright\arraybackslash}p{#1}}
\newcolumntype{C}[1]{>{\centering\arraybackslash}p{#1}}
\newcolumntype{R}[1]{>{\raggedleft\arraybackslash}p{#1}}

\begin{document}
\title{Improving Masked Autoencoders by Learning \\Where to Mask}
\titlerunning{Improving masked autoencoders by learning where to mask}
%
\author{Haijian Chen\inst{1}\and Wendong Zhang\inst{1}\and Yunbo Wang\inst{1}\textsuperscript{(}\Envelope\textsuperscript{)} \and Xiaokang Yang\inst{1}\textsuperscript{(}\Envelope\textsuperscript{)}}
\authorrunning{H.Chen et al}
\institute{Shanghai Jiao Tong University, Minhang District, Shanghai, China\\\email{\{higerchen, diergent, yunbow, xkyang\}@sjtu.edu.cn}}

\maketitle              
%
%
\begin{abstract}
   Masked image modeling is a promising self-supervised learning method for visual data. It is typically built upon image patches with \textbf{random} masks, which largely ignores the variation of information density between them. The question is: Is there a better masking strategy than random sampling and how can we learn it? We empirically study this problem and initially find that introducing object-centric priors in mask sampling can significantly improve the learned representations. Inspired by this observation, we present \textbf{AutoMAE}, a fully differentiable framework that uses Gumbel-Softmax to interlink an adversarially-trained mask generator and a mask-guided image modeling process. In this way, our approach can adaptively find patches with higher information density for different images, and further strike a balance between the information gain obtained from image reconstruction and its practical training difficulty. In our experiments, AutoMAE is shown to provide effective pretraining models on standard self-supervised benchmarks and downstream tasks.
\end{abstract}
\section{Introduction}



Inspired by the masked language modeling framework, \textit{masked autoencoder} (MAE) \cite{he2022masked,xie2022simmim}, also known as \textit{masked image modeling} (MIM), has become a promising self-supervised learning method based on Vision Transformer (ViT) \cite{dosovitskiy2021an}.
The general idea of this framework is to extract semantic representations by learning to maximize the log-likelihood function of unobserved patches for input images.


There are two problems that are coupled and crucial to the quality of the learned representations: \textit{First, how to determine the masks over the images? Second, how to learn from the masked images?} 
Previous literature has not been able to analyze the first problem sufficiently and satisfactorily in spite of its importance to the learning results.
Further, the solutions to ``\textit{where to mask}'' and ``\textit{how to learn from the masked data}'' have never been integrated into an end-to-end optimization scheme.

\begin{figure*}[t]
\begin{center}
\includegraphics[width=1.0\linewidth]{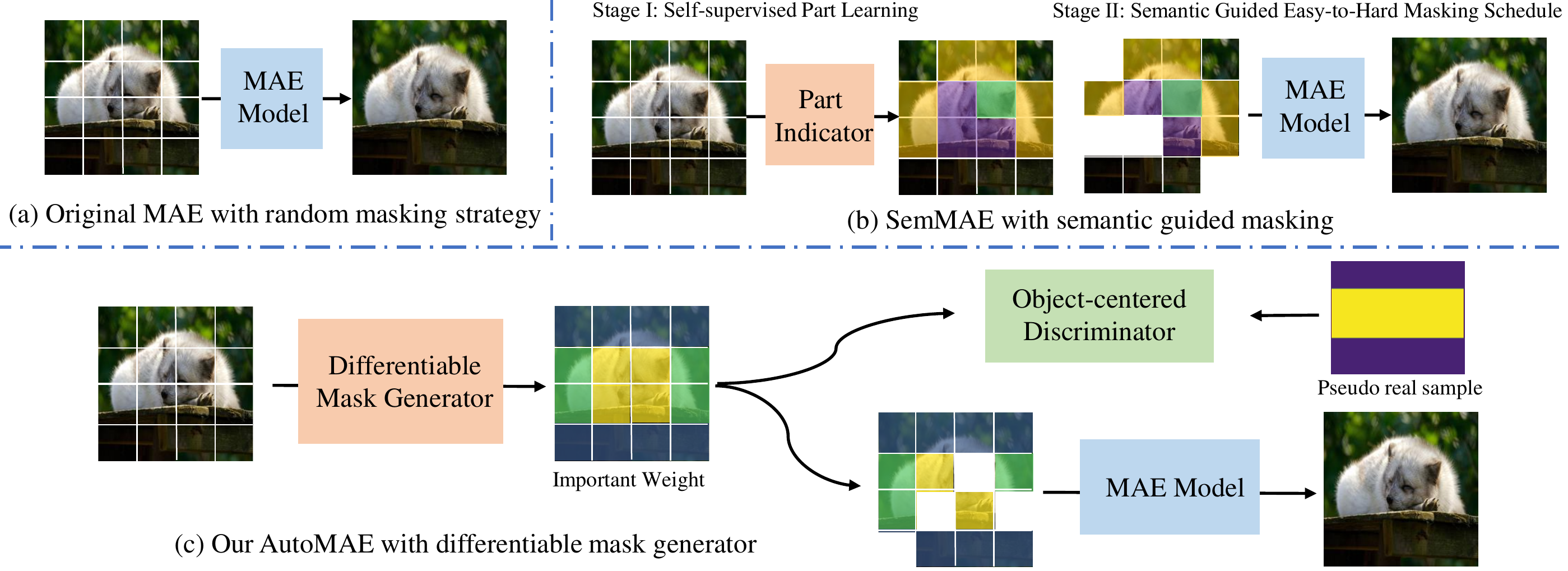}
\end{center}
\vspace{-15pt}
    \caption{Comparison of masking strategies. \textbf{(a)} The original MAE~\cite{he2022masked} randomly masks $70\%$ image patches with a uniform probability. \textbf{(b)} SemMAE~\cite{li2022semmae} uses a manually designed \textit{easy-to-hard} masking schedule guided by an independently-trained semantic part indicator. \textbf{(c)} AutoMAE is a fully differentiable framework that uses an adversarially-trained mask generator.
    }
\vspace{-10pt}
\label{fig:intro}
\end{figure*}



What should be a good masking strategy?
In general, the difficulty of pixel reconstruction is related to the information density of corresponding image patches. 
It is a paradox that if we mask a large number of ``hard'' patches (such as foreground objects), the model may fail to perceive the high-semantic areas; However, if we mask a large number of ``easy'' patches (such as the background), the self-supervised training task can be too simple for the model to learn effective representations.
In our preliminary experiments, we find that both scenarios result in the degeneration of the learned visual representations, as illustrated in Section~\ref{prelim-section}. We refer to such an empirical phenomenon as the \textit{patch selection dilemma} in MAE.



Therefore, we need to weigh between the informative and less-informative patches in the masking strategy, such that during pixel reconstruction, the model can reason about the relations between the missing semantics and the remainders. 
The random masking approach in the original MAE \cite{he2022masked} cannot adaptively perceive the variation of information density and applies a unified masking probability over the entire image. 
Following this line, SemMAE~\cite{li2022semmae} presents a two-stage MAE training scheme with a manually designed \textit{easy-to-hard} masking scheduler, guided by an independently trained semantic part indicator.
However, the existing approaches do not directly optimize the masking strategy, thus leading to sub-optimal solutions to constructing the pretext task.


In this paper, we propose AutoMAE, a pilot study of optimizing the masking strategy in a fully differentiable MAE framework.
As shown in Fig.~\ref{fig:intro}, we compare the key differences between AutoMAE and previous work. 
The key insights are two-folded:
First, it exploits an adversarially trained mask generator, in which the output masking probabilities are correlated with sample-specific information density across the patches, and are prone to be higher for ``hard-to-reconstruct'' regions in the foreground.
Second, it jointly optimizes the mask generator and the self-supervised ViT model through Gumbel-Softmax reparameterization. Intuitively, we may prevent the mask generator from overfitting the ``hard'' regions by back-propagating the reconstruction error of the unobserved patches through the entire model.




Our approach achieves competitive results in the linear probing and finetuning setups on ImageNet-1K~\cite{DBLP:conf/cvpr/DengDSLL009}.
Furthermore, it presents excellent transfer learning ability on CUB-Bird~\cite{Wah2011TheCB}, Stanford-Cars~\cite{DBLP:conf/iccvw/Krause0DF13}, iNaturalist 2019~\cite{van2018inaturalist}, COCO~\cite{DBLP:conf/eccv/LinMBHPRDZ14}, and ADE20K~\cite{zhou2019semantic} for fine-grained classification, detection, and segmentation, especially on small datasets with limited finetuning data. 

The main contributions of this paper are as follows:
\vspace{-5pt}
\begin{itemize}
    \item We demonstrate that the previous random sampling method is a suboptimal masking strategy in MIM, and the pretraining results can be significantly improved by slightly raising the masking rates of the informative foreground image patches.
    \item Motivated by these findings, we provide an early study of \textit{adversarial mask generation}, which incorporates simple structural priors into the adaptive masking rates.
    \item Unlike prior work, we propose to \textit{integrate mask generation and image reconstruction in a differentiable framework} using Gumbel-Softmax. It improves the pretraining results by striking a balance between the information gain through image reconstruction and its training difficulty.
\end{itemize}

\vspace{-5pt}
\section{Preliminaries}
\label{prelim-section}


\vspace{-5pt}
\subsection{Rethink the Mask Sampling Strategy in MAE}
\vspace{-3pt}

\paragraph{Revisiting MAE~\cite{he2022masked}.}

The learning process of MAE can be formalized as the following three steps. 
First, given an image $\boldsymbol{I} \in \mathbb{R}^{c\times h\times w}$ where $h$, $w$ and $c$ represent the image height, width, and channel numbers, respectively, MAE first divides the input image into $n=hw/p^2$ non-overlapping patches with spatial size equal to $p \times p$, where $p$ is the patch size. 
Then these divided patches are embedded by a linear projection layer and added with position embeddings to get the embedded tokens $\boldsymbol{Z} = \{ \boldsymbol{z}_i \}_{i=1}^{n}, \boldsymbol{z}_i\in \mathbb{R}^{d}$ where $d$ represents the token dimension. 
Next, in the encoding stage, MAE randomly selects only $25\%$ image tokens combined with a learnable \texttt{[CLS]} token $\boldsymbol{z}^{c}\in \mathbb{R}^{d}$ to form the input for the transformer-based encoder. These tokens are served as visible hints and processed by the encoder to obtain encoded tokens $\boldsymbol{Z}_{e}$. Given these embedded tokens, MAE starts the masked image modeling process which includes two stages: the encoding stage for visible token interactions and the decoding stage for removed content reconstruction. 
Finally, in the decoding stage, a shared mask token $\boldsymbol{z}^{m}\in \mathbb{R}^{d}$ is repeated and used to fill the places of previously dropped tokens. 
These repeated mask tokens combined with encoded tokens $\boldsymbol{Z}_{e}$ are further processed by the transformer-based decoder to reconstruct the pixel values for each dropped patch. 
The model is optimized by minimizing the reconstruction loss $\mathcal{L}_{\text{recon}}$ in dropped regions.

\myparagraph{Patch masking strategies.}
Among the above three steps, the masking strategy plays an important role in masked image modeling, which determines what kind of information will be learned. Although MAE adopts a high masking ratio, image patches are still dropped with equal probability. The random masking strategy ignores the difference in information density between patches and results in an ineffective learning process.

\myparagraph{Two-stage MAE.} 
To improve the random masking strategy, SemMAE~\cite{li2022semmae} exploits a two-stage framework that first learns the possible semantic parts without supervision and then uses the learned partition to determine which token should be masked. 
The high-quality part-level partition learned by SemMAE can be seen as local semantic guidance to help learn more informative image patches. 
However, SemMAE still needs a manually designed easy-to-hard masking schedule to regularize the learning process and the semantic number is fixed among different images.

\begin{figure}[t]
\centering
\includegraphics[width=0.75\linewidth]{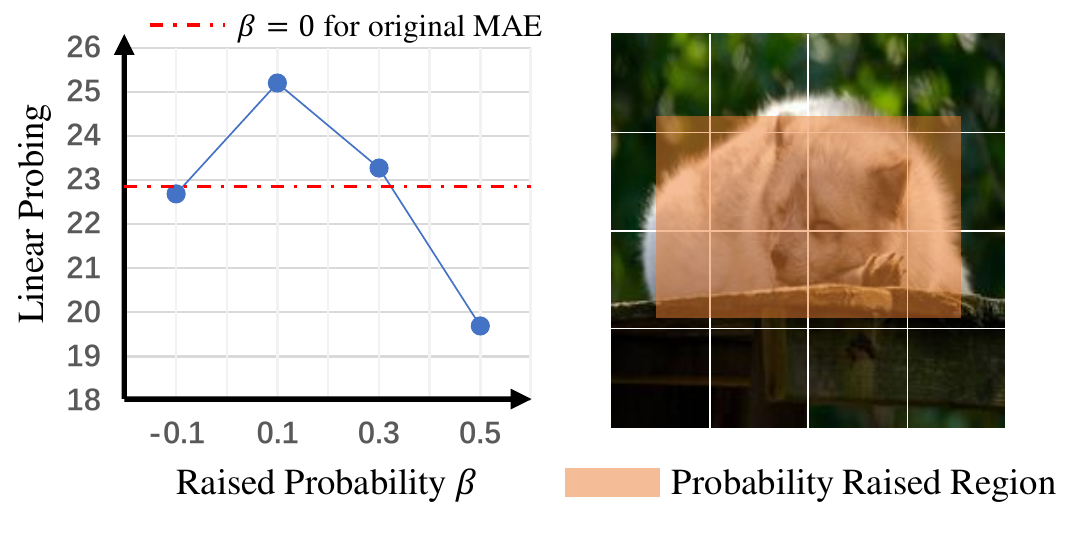}
\vspace{-15pt}
    \caption{Effects of raising the masking probability by $\beta$ on patches within the object bounding boxes. Models are trained on a subset ($10\%$) of the ImageNet dataset. The red dashed line represents the results of using the original random masking strategy.
    }
\vspace{-10pt}
\label{fig:preliminary}
\end{figure}

\vspace{-5pt}
\subsection{Key Findings: MAE with Prior Object Hints}
\label{sec:preliminary_observation}
\vspace{-3pt}

Unlike language modeling, the information density of images is more sparse and usually dominated by the objects that appear in the image.
Therefore we conduct a preliminary experiment to explore whether introducing the object location priors with high information density can help representation learning.
We randomly select a subset ($10\%$) of the ImageNet dataset~\cite{DBLP:conf/cvpr/DengDSLL009} and use the provided true object bounding boxes to indicate object locations. Instead of random masking, we manually raise the dropped probability of patches within the bounding box and use these samples for self-supervised pretraining. 
We experiment with different values and show the linear probing results in Fig.~\ref{fig:preliminary}, where $\beta$ represents the additionally raised probability for patches within the bounding box against those in the background.
We have two observations from Fig.~\ref{fig:preliminary}: 
\begin{itemize}
    \item Compared with random masking, slightly increasing the masking probability for patches within the bounding box significantly improves the linear probing results.
    \item Excessively raising the masking probability of patches within the bounding box may affect the validity of the pretext task and degrades the performance.
\end{itemize}
These results indicate that different image patches provide different effects on the learning of visual representations, and learning to reconstruct more foreground objects with higher information density can lead to better pretraining results. 
Nonetheless, \textit{we need to strike a balance between the information gain through masked image modeling and its training difficulty.}
Inspired by this, a natural solution is to integrate mask generation and image reconstruction into a fully differentiable framework. 
For different images, it can adaptively provide masking weights that reflect prior object hints and are friendly to masked image modeling, that is, \textit{we want to find patches with lower reconstruction difficulty within the obtained areas with higher information density.}
\vspace{-5pt}
\section{Method}
\vspace{-3pt}




\begin{figure*}[t]
    \centering
    \includegraphics[width=\textwidth]{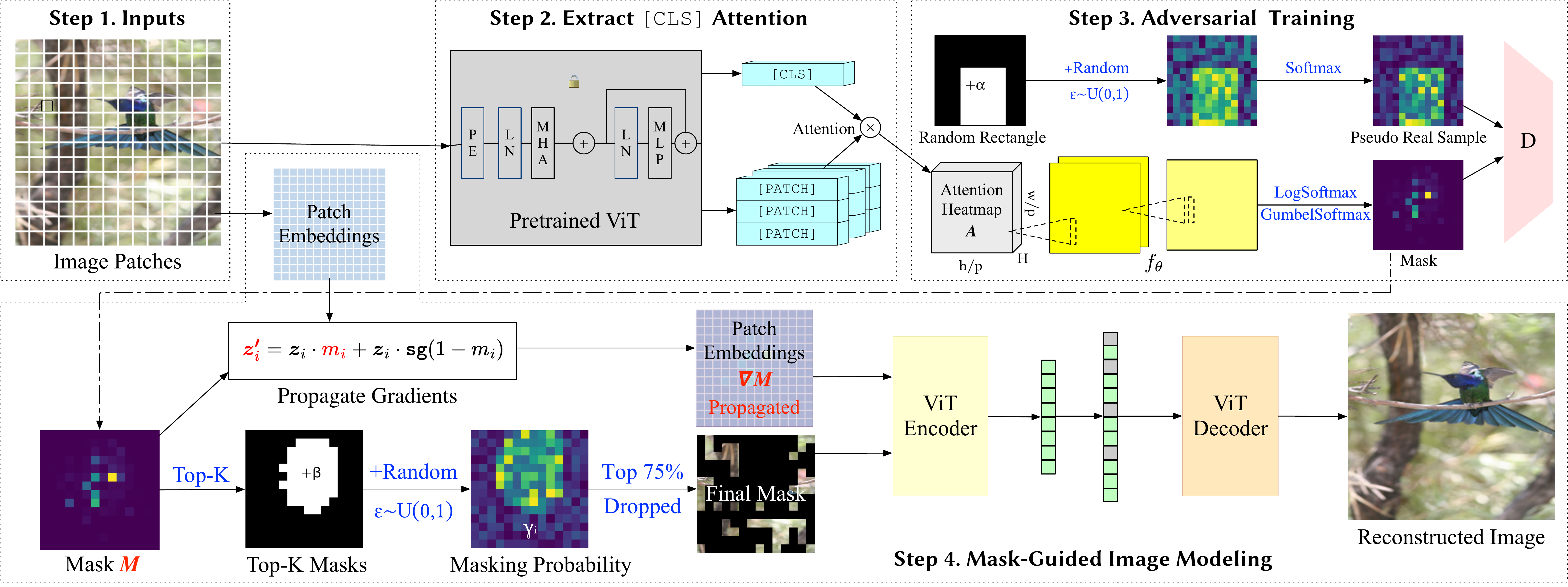}
    \vspace{-20pt}
    \caption{The end-to-end framework of AutoMAE, which is designed to tackle to \textit{patch selection} dilemma in a fully differentiable manner.}
    \label{fig:framework}
    \vspace{-10pt}
\end{figure*}


\subsection{General Design of AutoMAE}
\vspace{-3pt}
Motivated by our observations in Section~\ref{sec:preliminary_observation}, we design a novel self-supervised framework that can learn to select more informative foreground patches to mask, in which two key challenges need to be properly handled: First, how to guide the ViT model mining the informative patches without any explicit supervision? Second, how to control the difficulty of the pretext task via an adaptively learned mask strategy?

For the first challenge, we propose a differentiable mask generator $G$ to generate the mask image which contains the important weight for each image patch. The mask generator contains a pretrained ViT encoder (also in a self-supervised manner) and two trainable convolution layers. Given an input image, we use the pretrained ViT encoder to extract the multi-head attention maps from the last transformer block. The obtained attention maps contain indications of different semantic regions of the image and are further processed by the convolution layers for mask generation.
Since we cannot obtain explicit supervision to highlight the location of the foreground object, we introduce an object-centered adversarial training strategy to guide the mask generator in producing higher weights on patches within possible foreground objects.
Specifically, we introduce a discriminator and generate real mask images by manually raising the masking probabilities of a randomly sampled rectangle area. In this way, we minimize the distribution distance between the generated mask and the sample real mask, which helps the model find the informative foreground patches.

For the second challenge, we directly propagate the gradients from the masked autoencoder back to the mask generator and train these two modules synchronously. In other words, the reconstruction loss applied for the masked autoencoder also constrains the mask generator from generating too hard mask images. 
By this means, the masked autoencoder encourages the mask generator to produce high weights on patches that can easily infer the masking information, while the discriminator regularizes the mask generator and focuses more on possible foreground objects instead of the background.
These two branches cooperate with each other to facilitate the masked autoencoder to learn more representative patch relations.

\vspace{-8pt}
\subsection{Adversarially-Trained Mask Generator}
\vspace{-5pt}

The framework of AutoMAE is shown in Fig.~\ref{fig:framework}. 
The differentiable mask generator $G$ aims to generate sample-specific mask images to guide the token selection process in the training of the masked autoencoder, which mainly consists of a pretrained ViT encoder and two trainable convolutional layers.
We first exploit an MAE-pretrained~\cite{caron2021emerging} ViT encoder\footnote{Different self-supervised pretrained ViT models can be used for attention map extraction. Their effects are compared in our experiment. In the rest of this paper, we use the MAE-pretrained ViT-Base model without loss of generality.} to embed the image $\boldsymbol{I}\in \mathbb{R}^{c\times h\times w}$ and extract the multi-head self-attention map $\boldsymbol{A}\in \mathbb{R}^{H \times \frac{h}{p}\times \frac{w}{p}}$ between the query embedding of the \texttt{[CLS]} token and key embeddings of other patch tokens from the last transformer block, where $H$ represents the number of heads. 
Specifically, the calculation of the $i$-th attention map $\boldsymbol{A}_{i}\in \mathbb{R}^{1 \times \frac{h}{p}\times \frac{w}{p}}$ can be formalized as $ \boldsymbol{A}_{i} = \texttt{softmax}(\boldsymbol{q}_{i}^{c} \boldsymbol{K}_{i}^{\top} / \sqrt{d^\prime}), $
where $\boldsymbol{q}_{i}^{c} \in \mathbb{R}^{1 \times d^\prime}$ represents the query embedding of the \texttt{[CLS]} token, $\boldsymbol{K}_{i}\in \mathbb{R}^{\frac{hw}{p^{2}} \times d^\prime}$ represents the key embeddings of other patch tokens, and $d^\prime = d/H$ represents the embedding dim. Then we reshape the attention map with spatial size equal to $1 \times \frac{h}{p}\times \frac{w}{p}$. 
As mentioned in~\cite{caron2021emerging}, different attention maps in the last transformer block can attend to different semantic regions of the image. Therefore these attention maps can be served as informed initialization for further mask generation.
Besides, the parameters of the pretrained ViT encoder are frozen during the training process.
%
Given the multi-head self-attention map $\boldsymbol{A}$, we further exploit two convolutional layers with ReLU activation $f_{\theta}$ to embed these attention maps $ \boldsymbol{F} = f_{\theta}(\boldsymbol{A}), $ where $\boldsymbol{F}\in \mathbb{R}^{1 \times \frac{h}{p}\times \frac{w}{p}}$. We then sample the final mask image $\boldsymbol{M}\in \mathbb{R}^{1 \times \frac{h}{p}\times \frac{w}{p}}$ given the embedding feature $\boldsymbol{F}$:
\begin{equation}
    f_{i}^\prime = \log (\frac{\exp(f_{i})}{\sum_{k=1}^{hw/p}\exp(f_{k})}) + z, m_{i} = \frac{\exp(f_{i}^\prime)}{\sum_{k=1}^{hw/p}\exp(f_{k}^\prime)},
\label{eq:3}
\end{equation}
where $\log(softmax(\cdot))$ is used to stabilize the training process and $z$ represents a random noise sampled from a Gumbel distribution. The values in $\boldsymbol{M}$ are continuous and can be seen as the important weight of each image patch. We use an extra discriminator and the masked autoencoder itself to supervise the generation process. 



Although the attention map $\boldsymbol{A}$ can provide indications for different semantic regions, how to supervise the mask generator to produce a proper pretext task that can focus on foreground objects is still a challenging problem. 
To tackle this problem, we introduce an extra discriminator and design an object-centered adversarial training strategy to regularize mask generation.
The key insight of this approach is to generate pseudo mask images $\boldsymbol{M}^{p}$ as ``real'' samples which can imitate the difference between foreground and background patches, and use the adversarial training strategy to minimize the distribution shift between generated mask $\boldsymbol{M}$ and ``real'' samples $\boldsymbol{M}^{p}$.

The way to generate pseudo-mask images is quite simple. Given a zero-initialized image $\boldsymbol{M}^{p}\in \mathbb{R}^{1 \times \frac{h}{p}\times \frac{w}{p}}$, we first randomly sample a rectangle area within this image whose spatial size is between $20\% \sim 80\%$ of the whole image area. This sampled rectangle area is assumed to be the bounding box of one pseudo object.
Then we flatten the pseudo mask $\boldsymbol{M}^{p}=\{m_{i}^{p}\}_{i=1}^{n}$ and suppose the index set within the sample rectangle area is $\boldsymbol{X}$, the values in the pseudo mask are sampled as follows:
\begin{eqnarray}
    m_{i}^{p} = \epsilon+
    \begin{cases}
        \alpha& i \in \boldsymbol{X} \\
        0& i \notin \boldsymbol{X}
    \end{cases}
    \label{eq:4}
\end{eqnarray}
where $\epsilon$ represents a random noise sampled from a uniform distribution $\epsilon \sim U(0, 1)$, and $\alpha$ represents the additional important weight for patches within the rectangle, which is set to $0.5$ in our experiments. 
In this way, we generate the pseudo mask images where foreground patches contain higher weights compared with background patches. We then use the pseudo-mask images as a real sample for adversarial training with the loss function from LSGAN~\cite{DBLP:conf/iccv/MaoLXLWS17}:
\begin{equation}
\begin{aligned}
     \mathcal{L}_{\text{adv}} &= -\mathbb{E}_{M}  \left[\left(D(\boldsymbol{M}) - c \right)^{2}\right],\\
    \mathcal{L}^{D}_{\text{adv}} &= \mathbb{E}_{M^{p}} \left[ \left(D(\boldsymbol{M}^{p}) - b \right)^{2} \right] +
    \mathbb{E}_{M} \left[\left(D(\boldsymbol{M}) - a \right)^{2} \right], 
\end{aligned}
\label{eq:5}
\end{equation}
where $D$ is the discriminator and we set $a=-1,b =1, c=0$ as in previous work.

\vspace{-8pt}
\subsection{Mask-Guided Image Modeling}
\vspace{-3pt}

Given the sampled mask $\boldsymbol{M}$, we suggest that patches with higher weights are more likely to be foreground objects and more informative than patches with lower weights. 
%
%
Therefore, we raise the masking probabilities of these high-weighted patches and use low-weighted patches as visible hints in masked image modeling.
Specifically, we first sort the values in $\boldsymbol{M}=\{m_{1}, m_{2}, \cdots, m_{n}\}$ and then get the top $K$ index set $\boldsymbol{Y}$. The unnormalized masking probability of each token can be sampled as follows:
\begin{eqnarray}
    \vspace{-10pt}
    \gamma_{i} = \epsilon+
    \begin{cases}
        \beta& i \in \boldsymbol{Y} \\
        0& i \notin \boldsymbol{Y}
    \end{cases}
    \label{eq:6}
    \vspace{-10pt}
\end{eqnarray}
where $\epsilon$ represents a random noise sampled from a uniform distribution $\epsilon \sim U(0, 1)$, and $\beta$ represents the additionally raised probability for tokens with higher weights. In our experiment, we set $K=n/4$ and $\beta=0.5$ where $n$ represents the number of patch tokens. In this case, patches in set $\boldsymbol{Y}$ are supposed to be more informative than other patches, and models are supposed to pay more attention to these patches.

After that, we use the sampled probability to guide the token selection process where tokens with the top $75\%$ biggest probability are dropped and the remaining tokens with $25\%$ smallest probability are saved as visible hints for masked image modeling. 
In other words, we still follow the mask ratio used in the original masked autoencoder while using the sampled probabilities to enhance the learning of more important patches.

Particularly, we slightly change the formulation of the input patch tokens to allow the gradients from the masked autoencoder can be propagated back to the mask generator. Given the embedded patch tokens $\boldsymbol{Z} = \{ \boldsymbol{z}_i \}_{i=1}^{n}$ and the generated mask image $\boldsymbol{M}= \{m_i \}_{i=1}^{n}$, the new embedded patch tokens $\boldsymbol{Z^\prime}= \{ \boldsymbol{z^\prime}_i \}_{i=1}^{n}$ are obtained via $ \boldsymbol{z^\prime}_i = \boldsymbol{z}_i\cdot m_i + \boldsymbol{z}_i\cdot \texttt{sg}(1 - m_{i}), $
where $\texttt{sg}(\cdot)$ represents the stop gradient operation. We conduct token selection on these new embedded tokens $\boldsymbol{Z^\prime}$ and then perform the following encoding and decoding stages which are the same as the original masked autoencoder. In this way, we preserve the values of patch tokens while allowing the masked autoencoder to influence mask generation. Combined with the adversarial loss, the full objective function of the mask generator can be written as $ \mathcal{L}_\text{G} = \mathcal{L}_\text{recon} + \lambda\mathcal{L}_\text{adv} $ where we set $\lambda = 0.2$ by grid search.

\vspace{-8pt}
\section{Experiments}
\vspace{-5pt}

We evaluate AutoMAE in the linear probing and finetuning setups on ImageNet-1K~\cite{DBLP:conf/cvpr/DengDSLL009}, and perform model analyses on its smaller subsets. We also conduct transfer learning experiments on CUB-Bird~\cite{Wah2011TheCB}, Stanford-Cars~\cite{DBLP:conf/iccvw/Krause0DF13}, iNaturalist 2019~\cite{van2018inaturalist}, COCO~\cite{DBLP:conf/eccv/LinMBHPRDZ14}, and ADE20K~\cite{zhou2019semantic} for fine-grained classification, detection, and segmentation tasks.

We use the same ViT-Base architecture and a $16\times16$ patch size in all compared models. 
For AutoMAE, we typically use an $800$-epochs-pretrained MAE-Base (called ``MAE-800'') as the warmup model of the feature extractor in mask generation.
We freeze its parameters and train other parts of AutoMAE for $800$ epochs. 
We use SemMAE~\cite{li2022semmae} as our primary baseline as its mask generator also uses a warmup model\footnote{SemMAE~\cite{li2022semmae} uses an $800$-epochs iBOT model in the mask generator.}. 
We use the provided SemMAE pretrained model with a $16\times 16$ patch size.

\vspace{-8pt}
\subsection{Experiments on ImageNet-1K}
\label{sec:exp-full}
\vspace{-5pt}

When we have finished the preliminary model analyses on small subsets, we start to train AutoMAE on the full ImageNet-1K and analyze the experimental results.

\myparagraph{Linear probing.}
We use $3$ alternative models to initialize the feature extractor in the mask generator:
(1) the ViT-B network pretrained with MAE~\cite{he2022masked} (termed as MAE-800), (2) the ViT-B pretrained with iBOT~\cite{zhou2021ibot} (termed as iBOT-B), and (3) the ViT-S pretrained with DINO~\cite{caron2021emerging} (termed as DINO-S).
We use the LARS optimizer to train the model for $90$ epochs. The base learning rate is set to $0.1$, the momentum is $0.9$, and the batch size is $16{,}384$. We use random resized cropping and horizontal flipping for data augmentation. 
In Table \ref{tab:linear-probing}, AutoMAE consistently achieves better performance than the vanilla MAE model, regardless of the base models used for mask generation.


\begin{table}[t]
\begin{center}
\begin{tabular}{|L{2cm}|C{2cm}|C{3cm}|}
    \hline
    Model      & Warmup in $G$        & Acc-1(\%) \\ \hline
    MAE~\cite{he2022masked}               & N/A & 63.7    \\ 
    SemMAE~\cite{li2022semmae}          & iBOT-800  & 65.0    \\ 
    \hline
    AutoMAE & MAE-800  & \underline{66.7}    \\
    AutoMAE & iBOT-B \cite{zhou2021ibot} & \underline{66.7} \\
    AutoMAE & DINO-S \cite{caron2021emerging} & \textbf{68.8}    \\
     \hline
\end{tabular}
\end{center}
\vspace{-5pt}
\caption{Linear probing results of models pretrained on the full ImageNet for $800$ epochs. Regardless of the methods used in the mask generator to warm up its feature extractor, AutoMAE consistently achieves better performance.}
\label{tab:linear-probing}
\end{table}

\begin{table}[t]
\vspace{-10pt}
\begin{center}
\begin{tabular}{|L{2cm}|C{2cm}|C{1cm}|C{2cm}|}
    \hline
    Model   & Warmup in $G$         &  Ratio    & Acc-1(\%)  \\ \hline
    MAE~\cite{he2022masked}  & N/A           &  100    & 83.26      \\
    SemMAE~\cite{li2022semmae}  & iBOT-800         &  100    & \textbf{83.34}      \\
    AutoMAE & MAE-800 &  100    & \underline{83.32}      \\
    \hline
    MAE~\cite{he2022masked}  & N/A  &  30     & 78.95      \\ 
    SemMAE~\cite{li2022semmae} & MAE-800 &  30     & 79.00      \\
    AutoMAE & MAE-800 &  30     & \textbf{79.14}      \\ \hline    
\end{tabular}
\end{center}
\vspace{-5pt}
\caption{Finetuning results on ImageNet-1K. ``Ratio'' represents the percentage of the data we used for finetuning. All methods use ViT-Base as the encoder with a patch size of $16 \times 16$. }
\label{tab:finetuning}
\vspace{-20pt}
\end{table}

\myparagraph{Finetuning.}
In the finetuning stage, the base learning rate is set to $5\times 10^{-4}$ and the AdamW momentum is configured as $\beta_1, \beta_2 = 0.9, 0.999$. The effective batch size is $1{,}024$. 
Similar to previous literature, we use the RandAug technique \cite{cubuk2020randaugment} for data augmentation and employ layer-wise learning rate decay ($0.75$), Mixup \cite{DBLP:conf/iclr/ZhangCDL18} ($0.8$), CutMix \cite{DBLP:conf/iccv/YunHCOYC19} ($1.0$), and DropPath \cite{DBLP:conf/eccv/HuangSLSW16} ($0.1$) in the finetuning stage. All compared models are trained for 100 epochs in the same setting.
As shown in Table \ref{tab:finetuning}, our approach achieves competitive results to MAE and SemMAE. 
It is reasonable that SemMAE has a slightly better performance on full-set finetuning, as it uses an iBOT-based mask generator which is pretrained by contrastive learning. 
It is also worth noting that SemMAE with a $16\times16$ patch size obtains similar finetuning results, indicating that the non-linear features for big tokens in MAE are strong enough for image classification. 
Furthermore, we perform an additional experiment to finetune the models on the $30\%$ subset. 
As shown in Table \ref{tab:finetuning}, AutoMAE outperforms MAE and SemMAE by large margins, suggesting that it is more effective in scenarios with limited data.


\vspace{-8pt}
\subsection{Downstream Tasks}
\label{sec:down}
\vspace{-5pt}

\paragraph{Fine-grained image classification.} 
We first finetune the models on fine-grained image classification datasets, including CUB-Bird~\cite{Wah2011TheCB}, Stanford-Cars~\cite{DBLP:conf/iccvw/Krause0DF13}, and iNaturalist 2019~\cite{van2018inaturalist}. 
As shown in Table~\ref{tab:finegraind_cls}, AutoMAE achieves the best performance on all datasets. 
Similarly, we experiment with random-sampled subsets ($50\%$ and $10\%$), where AutoMAE shows more significant improvements compared with other models. 
These results demonstrate the transfer learning ability of our method to small datasets.

\begin{table}[t]
\begin{center}
\begin{tabular}{|L{2cm}|C{1.5cm}|C{1.5cm}|C{1.5cm}|C{1.5cm}|}
\hline
Method            &Ratio(\%) & CUB & Cars & iNat-19 \\ \hline
MAE~\cite{he2022masked}              &100 & 83.3 & 92.7 & 79.50         \\
SemMAE~\cite{li2022semmae}           &100 & 82.1 & 92.4  & 79.60        \\
AutoMAE         &100 & \textbf{83.7} & \textbf{93.1} & \textbf{79.93}         \\ 
\hline
MAE~\cite{he2022masked}              &50 & 70.6 & 84.2 & 73.76           \\
SemMAE~\cite{li2022semmae}           &50 & 70.6 & 82.1 & 73.47     \\
AutoMAE         &50 & \textbf{73.4} & \textbf{84.6}  & \textbf{74.03}         \\ 
\hline
MAE~\cite{he2022masked}     &10   & - & -      & 49.14         \\
SemMAE~\cite{li2022semmae}  &10   & - & -    & 48.65     \\
AutoMAE                     &10   & - & -    & \textbf{50.13}  \\ \hline
\end{tabular}
\end{center}
\vspace{-5pt}
\caption{Fine-grained classification results. All methods exploit ViT-Base as the encoder with a patch size equal to $16 \times 16$. ``Ratio'' is the percentage of data used in fine-tuning. Ratio 10\% for CUB and Cars is omitted because the two datasets are not large enough.}
\label{tab:finegraind_cls}
\end{table}

\begin{table}[t!]
\vspace{-10pt}
\begin{center}
\begin{tabular}{|L{2cm}|C{3cm}|C{1.5cm}|C{1.5cm}|}
\hline
Method            & Pretraining Epochs & AP\textsuperscript{box} & AP\textsuperscript{mask} \\ \hline
BEiT*~\cite{DBLP:conf/iclr/Bao0PW22}              &800  &49.8      &44.4 \\
MAE*~\cite{he2022masked}               &1600 & 50.3     & 44.9      \\
AutoMAE          &800 & \textbf{50.5}     & \textbf{45.0}      \\ \hline
\end{tabular} 
\end{center}
\vspace{-5pt}
\caption{Results of object detection and instance segmentation on the COCO dataset. * indicates results from the original literature.}
\label{tab:detect}
\vspace{-20pt}
\end{table}

\begin{table}[t]
\begin{center}
\begin{tabular}{|L{4cm}|C{2cm}|C{2cm}|}
\hline
Model    & Ratio(\%) & mIoU \\ 
\hline
Supervised Pretraining & 100 &45.3 \\
\hline
BEiT*~\cite{DBLP:conf/iclr/Bao0PW22} & 100              &45.8 \\
MAE~\cite{he2022masked}    & 100           & 46.1 \\
SemMAE~\cite{li2022semmae} & 100          & 46.3 \\
AutoMAE & 100 & \textbf{46.4}  \\ 
\hline
MAE~\cite{he2022masked} &   50           & 41.7 \\
SemMAE~\cite{li2022semmae} & 50          & 41.9 \\
AutoMAE & 50 & \textbf{42.4} \\ \hline
\end{tabular}
\end{center}
\vspace{-5pt}
\caption{Results of semantic segmentation on ADE20K.}
\label{tab:semantic segmentation}
\vspace{-15pt}
\end{table}


\myparagraph{Object detection and instance segmentation.} 
Table~\ref{tab:detect} gives the detection and instance segmentation results on COCO.
We follow the previous work~\cite{DBLP:journals/corr/abs-2111-11429} to finetune the Mask-RCNN~\cite{DBLP:conf/iccv/HeGDG17} with a ViT-based FPN backbone. 
Due to the memory limit, we set the total batch size to $32$, half of the original value, and set the learning rate to $4\times 10^{-5}$. 
We finetune AutoMAE for $100$ epochs which is the same as the compared models. 
It shows that even though AutoMAE is pretrained for half epochs ($800$), it still outperforms the original MAE by $0.2$ in $\text{AP}^\text{box}$ and $0.1$ in $\text{AP}^\text{mask}$.

\myparagraph{Semantic segmentation.}
We use the ADE20K dataset for semantic segmentation, which contains $150$ semantic categories and $25$K images.
Similar to SemMAE~\cite{zhou2019semantic}, we use UperNet~\cite{xiao2018unified} as the network backbone to finetune the pretrained ViT-B network for $160$K iterations on ADE20K. 
%
The learning rate is set to $5\times10^{-5}$ and the batch size is $8$. We also finetune our pretrained model for $100$ epochs which is the same as other compared methods.
As shown in Table \ref{tab:semantic segmentation}, AutoMAE outperforms MAE by $0.3$ mIoU and outperforms SemMAE by $0.1$ mIoU.
We then use fewer data ($50\%$) in the finetuning stage. 
We observe that AutoMAE achieves a more significant advantage over MAE and SemMAE in this case, increasing mIoU results by $0.7$ and $0.5$, respectively.

\vspace{-8pt}
\subsection{Mask Visualization}
\begin{figure}[t!]
\begin{center}
\includegraphics[width=0.8\linewidth]{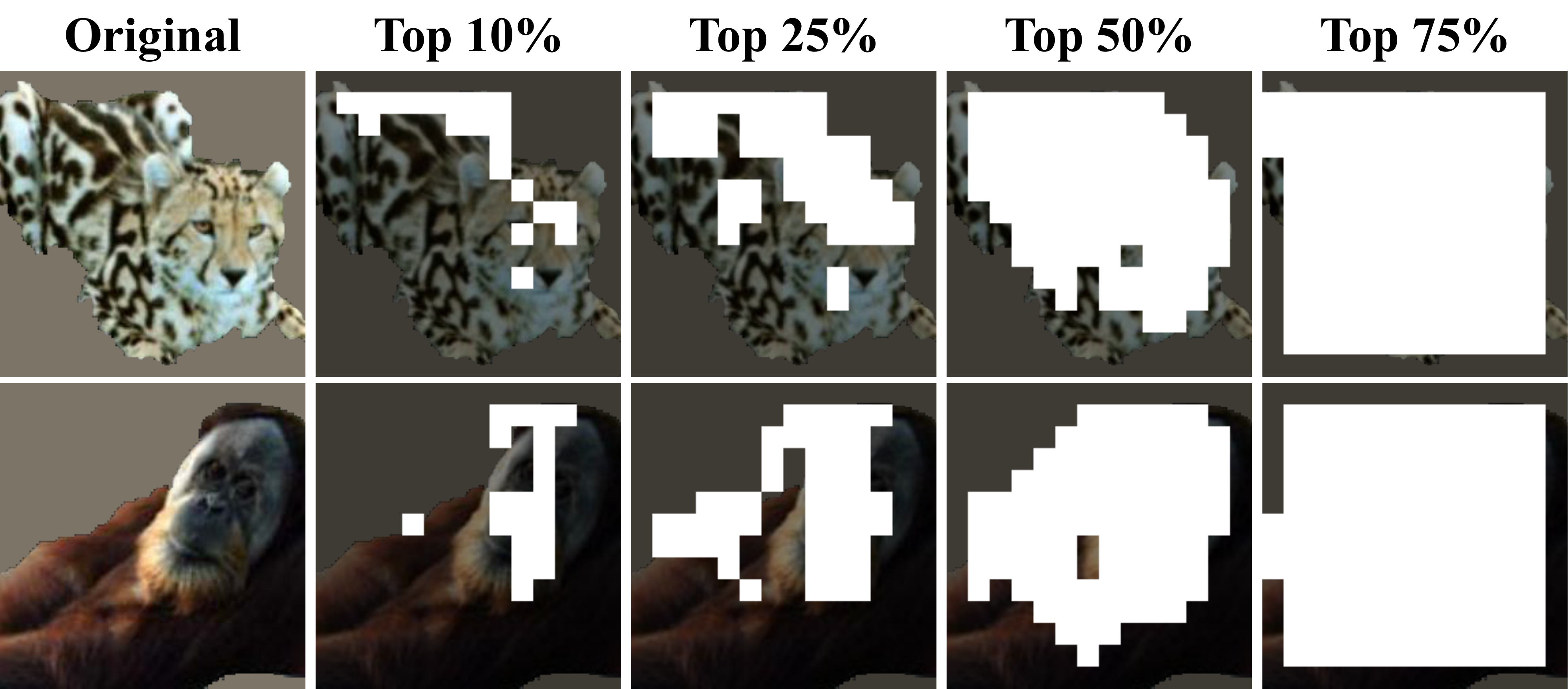}
\end{center}
\vspace{-15pt}
    \caption{High-weighted masks on \textbf{ImageNet-9} produced by the mask generator. The highlighted areas are obtained from the mask before adding random noise by $K$-largest values. 
    }
\label{fig:IN9}
\vspace{-10pt}
\end{figure}

\vspace{-5pt}
To verify whether the mask generator can perceive meaningful areas, we conduct experiments on ImageNet-9~\cite{xiao2021noise}, a dataset with foreground objects only. As shown in Fig.~\ref{fig:IN9}, the most high-weighted masks in AutoMAE mainly focus on meaningful patches, such as the back and eyes of the jaguar, and gradually involve less significant patches as the masking ratio increases. More results are provided in supplementary materials.

\vspace{-8pt}
\section{Related Work}
\vspace{-5pt}

The majority of recent studies of self-supervised learning for visual representations of deep networks can be roughly grouped into two categories, \textit{i.e.}, contrastive learning and masked image modeling. 

\myparagraph{Contrastive learning.}
Contrastive learning methods typically distinguish positive samples from other samples. Early work like MoCo~\cite{he2020momentum} and SimCLR~\cite{chen2020simple} focuses on distinguishing the negative pairs, while BYOL \cite{grill2020bootstrap}, SimSiam~\cite{chen2021exploring} and DINO~\cite{caron2021emerging} eliminates the need for negative pairs. DINO is trained by self-distillation, and it observes that ViT features contain explicit information about the semantic segmentation of an image. 

\myparagraph{Masked image modeling.} 
Recent studies in masked image modeling (MIM)~\cite{DBLP:journals/corr/abs-2205-09113,DBLP:journals/corr/abs-2203-12602,dosovitskiy2021an,DBLP:conf/iclr/Bao0PW22,chen2020generative} are largely inspired by the success of BERT \cite{DBLP:conf/naacl/DevlinCLT19}. 
According to different regression targets, MIM methods can be divided into two groups: pixel-level reconstruction and feature-level prediction. 
Feature-level prediction usually requires an external model called ``Tokenizer'' to generate reconstruction targets. BEiT~\cite{DBLP:conf/iclr/Bao0PW22} and PeCo~\cite{ramesh2021zero} are pretrained to predict visual tokens generated by discrete variational autoencoder (dVAE)~\cite{ramesh2021zero}. 
iBOT \cite{zhou2021ibot} presents an online tokenizer as the teacher network to produce the target and perform self-distillation. 
On the other hand, pixel-level reconstruction methods directly use raw pixel values as the regression targets. MAE \cite{he2022masked} and SimMIM \cite{xie2022simmim} show that masking a high ratio of patches and directly predicting RGB values can achieve BEiT-level performance. 
The group of work most relevant to AutoMAE includes methods that also improve masking strategies in MIM.
SemMAE~\cite{li2022semmae} adopts a two-stage framework that first learns a part indicator and then fixes it in the MIM training stage. AttMask~\cite{10.1007/978-3-031-20056-4_18} presents an independently-trained self-attention module to mask high-attended patches, while MST~\cite{li2021mst} proposes to mask low-attended patches.
\textit{In the above discussions, we empirically show the superiority of the proposed differentiable framework over two-stage training.}
ADIOS~\cite{shi2022adversarial} combines an adversarial MIM with contrastive learning.
\textit{In contrast, AutoMAE does not rely on contrastive learning but rather introduces prior hints through adversarial training. Furthermore, AutoMAE does not require expensive multiple forward passes in a single training iteration.}
The experiments of ADIOS are mainly conducted on small datasets such as STL10 and a downsized version of ImageNet-100 \cite{tian2020contrastive}, which supports the validity of our model analysis setup on the ImageNet subsets. 

\vspace{-2mm}
\section{Conclusion}
In this paper, we focused on improving the masking strategy that was shown to play an important role in the masked image modeling framework. We illustrated that introducing object-centric priors to guide the masking strategy can significantly improve the learned visual representations. Starting from this point, we proposed AutoMAE, which integrates a differentiable mask generator to provide the sample-specific categorical distribution of masking probabilities across all image patches. The mask generator is jointly trained with the ViT model with an adversarial learning constraint. We validated the effectiveness of AutoMAE on different benchmarks and downstream tasks.

\section*{Acknowledgments}

This work was supported by the National Natural Science Foundation of China (Grant No. 62250062, 62106144), the Shanghai Municipal Science and Technology Major Project (Grant No. 2021SHZDZX0102), the Fundamental Research Funds for the Central Universities, and the Shanghai Sailing Program (Grant No. 21Z510202133).

\bibliographystyle{splncs04}
\bibliography{egbib}

\newpage

\begin{center}
\textbf{\large Supplementary Materials}
\end{center}

\setcounter{section}{0}
\setcounter{equation}{0}
\setcounter{figure}{0}
\setcounter{table}{0}

\renewcommand{\theequation}{S\arabic{equation}}
\renewcommand{\thefigure}{S\arabic{figure}}
\renewcommand{\thesection}{S\arabic{section}}
\renewcommand{\thetable}{S\arabic{table}}
\section{Model Analyses}

We conduct model analyses in the linear probing setup by pretraining the models on smaller subsets of ImageNet-1K.
%
%
Specifically, we randomly sample $10\% \sim 30\%$ images and perform the pretraining for $400$ epochs. For linear probing, we finetune the model for $90$ epochs and evaluate it on the \textbf{full} ImageNet-1K validation set. 



\subsection{Mask Generator with A Pretrained ViT} 
In Table \ref{tab:ablations}, we compare the use of different self-supervised warmup methods (MAE-800 vs. MAE-400-Small) for the ViT-Base feature extractor in the mask generator. 
The latter one is \emph{pretrained on the $30\%$ subset for $400$ epochs}. 
It achieves a comparable Top-1 accuracy ($53.57\%$) with the one with the MAE-800 backbone ($53.58\%$), which is significantly higher ($+2.89$) than its baseline MAE ($50.50\%$) and the best MAE model with ground-truth bounding box guidance (as described in \ref{sec:preliminary_observation} of the paper). 
These results indicate that the main improvement in the performance of AutoMAE comes from the differentiable mask sampling method, as the mask generator is effective even with a weak feature extractor.

\begin{table}[h]
\begin{center}
\vspace{-10pt}
\begin{tabular}{|L{3cm}|L{3cm}|C{2cm}|}
\hline
Model  & Warmup in $G$  & Acc-1(\%)  \\ \hline
MAE~\cite{he2022masked}   & N/A & 50.50 \\
| w/ bounding box & N/A & 52.52 \\
\hline
AutoMAE & MAE-400-Small & 53.57 \\
AutoMAE  & MAE-800  & 53.58 \\
| Two-Stage & MAE-800 & 52.57 \\
| $G:$ MAE-Max   & MAE-800 & 50.85 \\
AutoMAE & DINO-B     & 53.20 \\
| $G:$ DINO-Max  & DINO-B & 51.52 \\ \hline
\end{tabular}
\end{center}
\vspace{-5pt}
\caption{Linear probing accuracy of models pretrained on a $30\%$ subset of ImageNet-1K for $400$ epochs. All models are on the full validation set of ImageNet-1K. We use different ViT backbones with frozen parameters in the mask generator.}
\label{tab:ablations}
\vspace{-40pt}
\end{table}

\begin{table}[t]
\begin{center}
\begin{tabular}{|L{3cm}|L{3cm}|C{2cm}|}
\hline
Model & Warmup in $G$    & Acc-1(\%)  \\ \hline
MAE~\cite{he2022masked}  & N/A     & 23.07 \\ 
AutoMAE & MAE-800   &\textbf{26.84}  \\
\hline
AutoMAE & From-Scratch    & \underline{26.00} \\ 
| w/ EMA $m=0.9$ & From-Scratch& 25.92 \\ 
| w/ EMA $m=0.99$ & From-Scratch& 25.45 \\ \hline
\end{tabular}
\end{center}
\vspace{-5pt}
\caption{Linear probing of models pretrained on $10\%$ ImageNet for $400$ epochs. AutoMAE outperforms the original MAE pertaining even with a randomly initialized mask generator.
}
\label{tab:ablations-supplementary}
\end{table}

\begin{table}[t!]
\begin{center}
\begin{tabular}{|L{3cm}|C{1cm}|C{1cm}|C{1cm}|}
\hline
$\beta$             & 0.2   & 0.5   & 1.0 \\ \hline
Linear probing     & 52.04 & 53.58 &  53.43  \\ \hline
\end{tabular}
\end{center}
\vspace{-5pt}
\caption{Linear probing accuracy with different margin values.}
\label{tab:beta_search}
\vspace{-15pt}
\end{table}


\subsection{Mask Generator without Pretraining} 
In most of the experiments, the mask generator of AutoMAE relies on a pretrained ViT. We here consider removing such a requirement by using a train-from-scratch ViT backbone. The simplest way is to \textbf{reuse} the same ViT model for both mask generation and mask reconstruction.
As shown in Table~\ref{tab:ablations-supplementary}, this model (``From-Scratch'') remarkably outperforms the vanilla MAE model ($26.00\%$ vs $23.07\%$), both pretrained on the $10\%$ ImageNet subset.
It suggests that AutoMAE is effective even without any pretrained models.
We further perform the exponential moving average (EMA) to synchronize the ViT used for mask generation ($k$) and the one used for image reconstruction ($q$). We have $\theta_k = m \times \theta_k + (1-m) \times \theta_q$. 
Results are shown in Table~\ref{tab:ablations-supplementary}.

\subsection{Joint Training} 
We conduct an ablation study of AutoMAE by separating it into two independent learning processes, i.e., mask generation and mask-guided image modeling.
In Table \ref{tab:ablations}, the ``Two-stage'' model outperforms MAE ($52.57\%$ vs $50.50\%$) but underperforms the joint training version ($53.58\%$), which validates the effectiveness of joint training in AutoMAE.
In Table 1 of the paper (\textit{full-set pretraining}), we compare AutoMAE with SemMAE, a two-stage pretraining pipeline with an independent mask learning stage. It further supports the superiority of joint training over two-stage training ($66.7\%$ vs. $65.0\%$).


\subsection{Adversarial Mask Generation} As DINO \cite{caron2021emerging} suggests that the last self-attention layer in ViT blocks can provide meaningful masks, we remove the adversarial training loss and replace the two-layer CNN in the mask generator (after the feature extractor) with a simple $\mathrm{maximum}$ operator over the channel axis. 
From Table \ref{tab:ablations}, although these models (``MAE-Max \& DINO-Max'') still outperform the original MAE, they obtain lower accuracy than corresponding AutoMAE models with MAE-800 and DINO-B.
It indicates that the mask generator does learn a better masking function than simply taking the maximum attention values.

\subsection{Sensitivity Analysis on $\beta$} In Fig. 2 in the main manuscript, we show how the margin value of the masking probability $\beta$ affects the linear probing performance with bounding box guidance. Here, as shown in Table \ref{tab:beta_search}, we conduct a grid search over $\beta=\{0.2, 0.5, 1.0\}$ on the ImageNet subset and find that $\beta=0.5$ performs best.

\section{Implementation Details}
\label{pre-train-section}

\subsection{Model Architecture}

\begin{figure}[t]
\begin{center}
\includegraphics[width=0.9\linewidth]{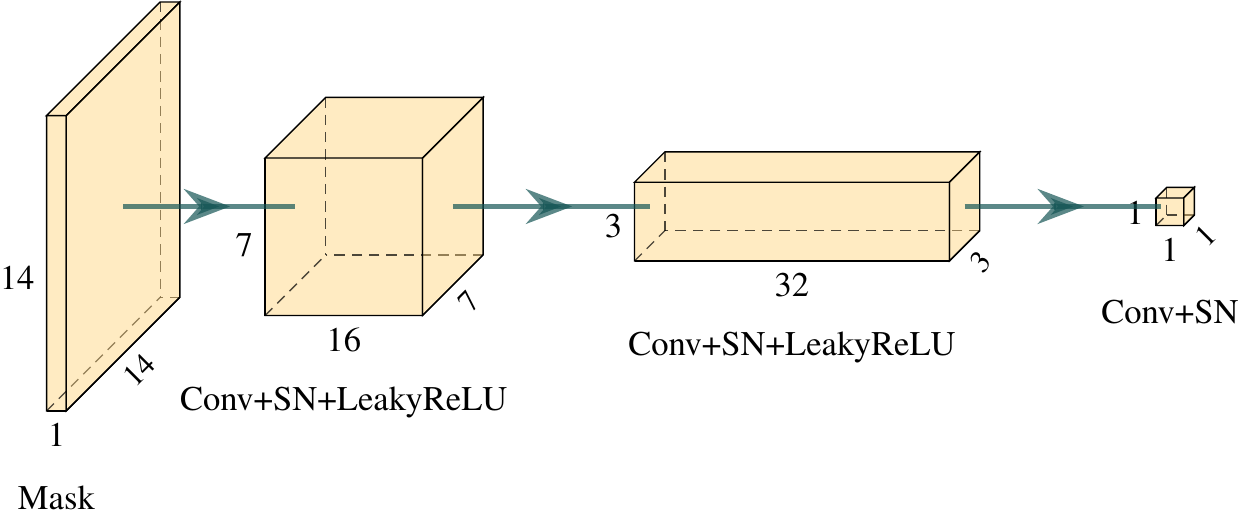}
\vspace{-10pt}
\caption{The architecture of the mask discriminator.}
\label{fig:discriminator}
\end{center}
\vspace{-5pt}
\end{figure}

In adversarial training, we use a three-layer CNN as the discriminator. The first two convolutional layers have a $4\times4$ kernel size, a stride of $2$, and a zero-padding length of $1$. They are followed by a Spectral Normalization~\cite{DBLP:journals/corr/YoshidaM17} layer and a LeakyReLU activation. The LeakyReLU has a negative slope of $0.2$. The last convolutional layer has a filter size of $3\times3$ and is also followed by a spectral normalization layer. As shown in Fig.~\ref{fig:discriminator}, the spatial size of feature maps is $14\rightarrow 7\rightarrow 3 \rightarrow 1$ and the number of channels is $1\rightarrow 16\rightarrow 32\rightarrow 1$.


\subsection{Training Details} We adopt the AdamW optimizer \cite{DBLP:conf/iclr/LoshchilovH19} for pretraining the MAE model and the mask generator in our AutoMAE. Following previous work~\cite{he2022masked}, the base learning rate is set to $1.5\times 10^{-4}$, the momentum is configured as $\beta_1 = 0.9, \beta_2 = 0.95$, and the batch size is $4{,}096$. We use the linear learning rate schedule: \textit{lr} = \textit{base\_lr} $\times$ \textit{batchsize} $/ 256$. The weight decay is $0.05$. The learning rate is scheduled by cosine decay with a warm-up phase of $40$ epochs and the total number of pretraining epochs is set to $800$. We use random resized cropping and random horizontal flipping for data augmentation. 
For the discriminator, we use the Adam optimizer without the weight decay. Other hyper-parameters stay the same. The mask generator uses a margin of $\alpha = 0.5$ to highlight the randomly sampled foreground area, and the loss weight $\lambda$ is set to $0.2$ to balance generator loss with reconstruction loss.  

\subsection{Hyperparameters Tuning}
In Eq.(4) and Eq.(6), we set $\alpha = \beta$ as they have similar effects. In Eq.(5), the values of $a,b,c$ are directly taken from LSGAN~\cite{DBLP:conf/iccv/MaoLXLWS17}. We only tune $\beta$ and the loss weight $\lambda$ on the ImageNet subset ($30\%$) and reuse the obtained values in other datasets/setups.

\begin{table}[t]
\begin{center}
\begin{tabular}{|l|c|c|C{3.5cm}|c|}
\hline
Method & Time (@800 epochs) & FLOPs & Hyper-parameters to tune & Better results   \\
\hline
SemMAE~\cite{li2022semmae} & 6 days & Unknown & Partitions, Blur kernel,  $\gamma$, Patch size, $\lambda$ & 1/7 setups\\ 
\hline
AttMask \cite{10.1007/978-3-031-20056-4_18} & $\textgreater$ 8 days & 37.03G & \#Attn layers, $p$, $r$, $s$, $\lambda$ (see 
\cite{10.1007/978-3-031-20056-4_18} for details) & --\\ 
\hline
AutoMAE (ours)  & 7 days & 35.72G & $\alpha=\beta$, $\lambda$ & 6/7 setups\\ 
\hline 
\end{tabular}
\end{center}

\caption{
AutoMAE performs better \textbf{in $6/7$ setups}: Linear probing, $30\%$ finetuning, $100\% \& 50\%$ fine-grained classification, detection, and segmentation. It is worth noting that: (1) The mask generators in SemMAE and AutoMAE are both based on the pretrained ViT encoder; (2) AutoMAE has fewer hyperparameters to tune than SemMAE and AttMask.
}
\vspace{-10pt}
\label{tab:compare}
\end{table}

\subsection{Training Cost} We use a patch size of $16\times 16$ for all experiments. It takes $26$ hours to pretrain an AutoMAE model on $30\%$ of ImageNet with $8$ NVIDIA RTX 3080Ti cards for $400$ epochs, and $7$ days to pretrain an AutoMAE model on the full ImageNet dataset with $8$ NVIDIA A100 cards for $800$ epochs. Admittedly, MAEs commonly benefit from a smaller $8\times8$ patch size. However, we cannot afford its training cost, \textit{e.g.}, it takes about $4$ weeks for $800$ epochs on the full ImageNet-1K. We want this paper to focus more on the proposed training scheme than on hyper-parameter tuning. Therefore, we follow the standard $16\times16$ setup~\cite{he2020momentum,zhou2021ibot}, which does not affect the main conclusions when compared to SemMAE.
As shown in Table~\ref{tab:compare}, AutoMAE has comparable time cost to recent advances in self-supervised pre-training \cite{10.1007/978-3-031-20056-4_18,li2022semmae} based on the same hardware ($8\times$A100 GPUs) for $800$ epochs. The full model has a similar number of FLOPs to existing work.

\section{Visualization}
\label{sec:3}

In this section, we will explain the details and show more samples from ImageNet-1K~\cite{DBLP:conf/cvpr/DengDSLL009} during the pretraining process of AutoMAE.

\begin{figure*}[!t]
    \centering
    \includegraphics[width=\textwidth]{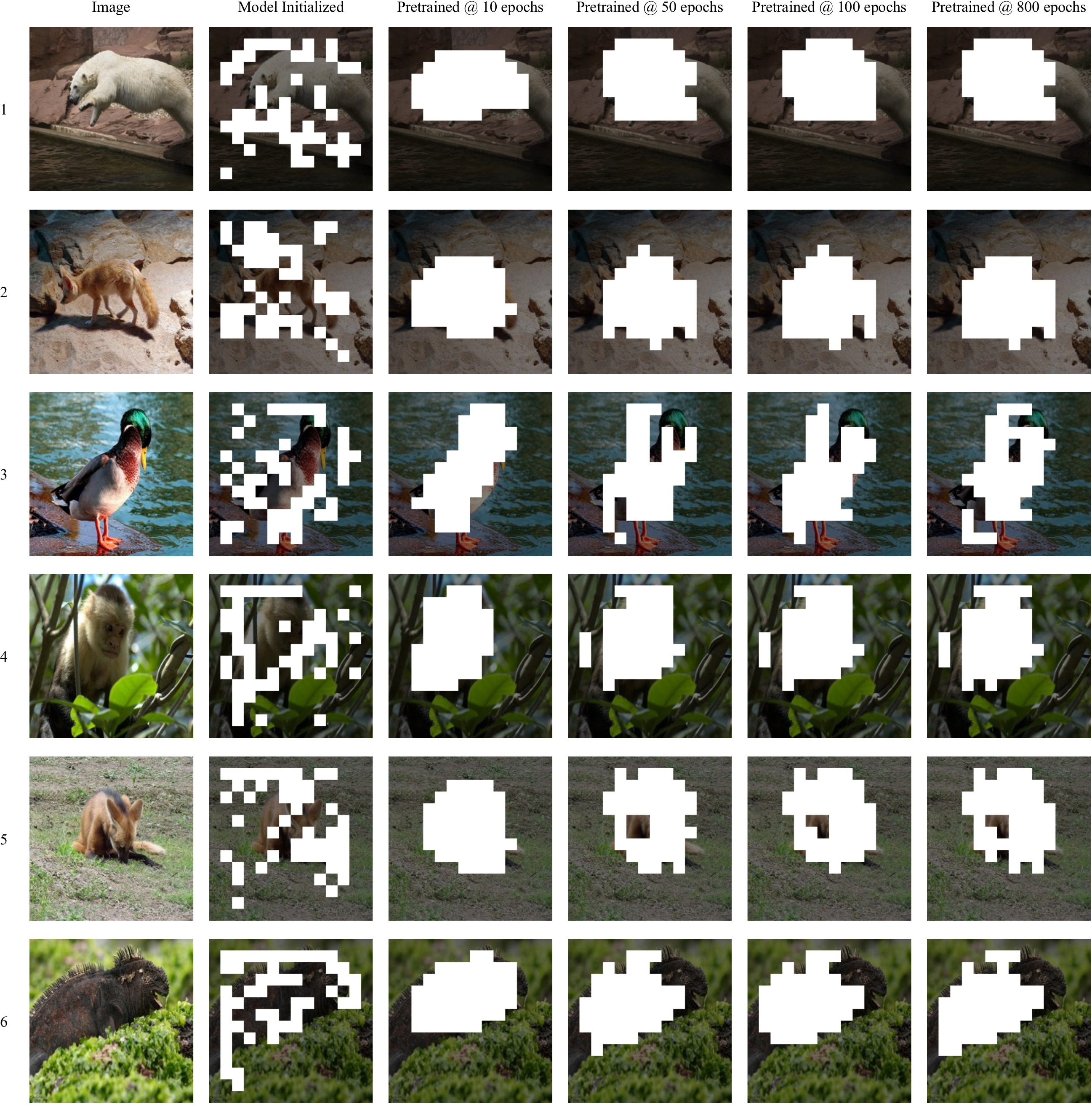}
    \caption{Masked patches by AutoMAE during the pretraining for $800$ epochs. White patches are the output of $\log(\mathrm{softmax}(\cdot))$ with the top $25\%$ probabilities.}
    \label{fig:masked1}
\end{figure*}

\begin{figure*}[!t]
    \begin{center}
        \includegraphics[width=\textwidth]{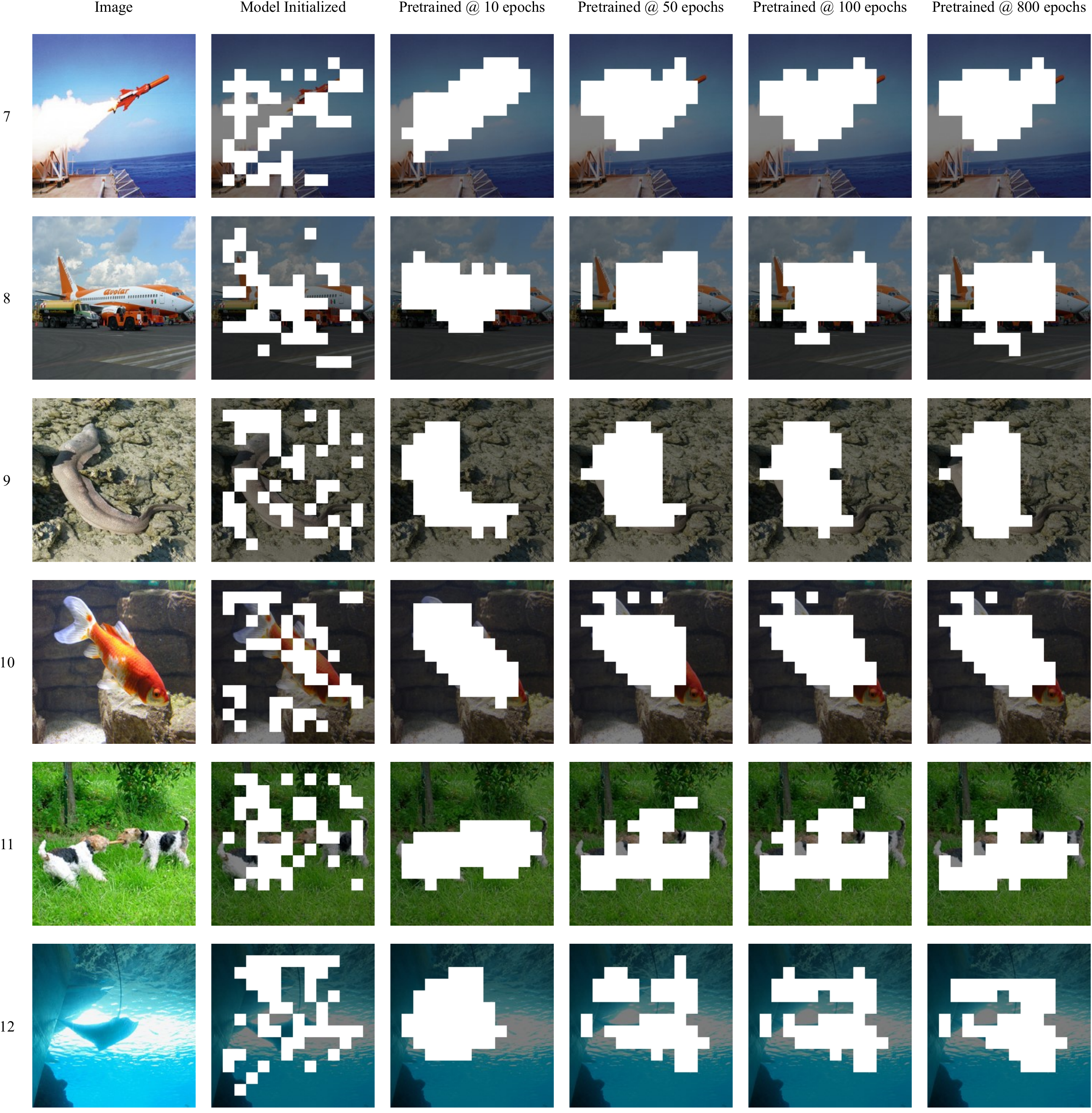}
        \caption{Masked patches by AutoMAE during the pretraining for 800 epochs. White patches are the output of $\log(\mathrm{softmax}(\cdot))$ with the top $25\%$ probabilities.}
    \label{fig:masked2}
    \end{center}
\end{figure*}

\begin{figure*}[!t]
    \begin{center}
        \includegraphics[width=\textwidth]{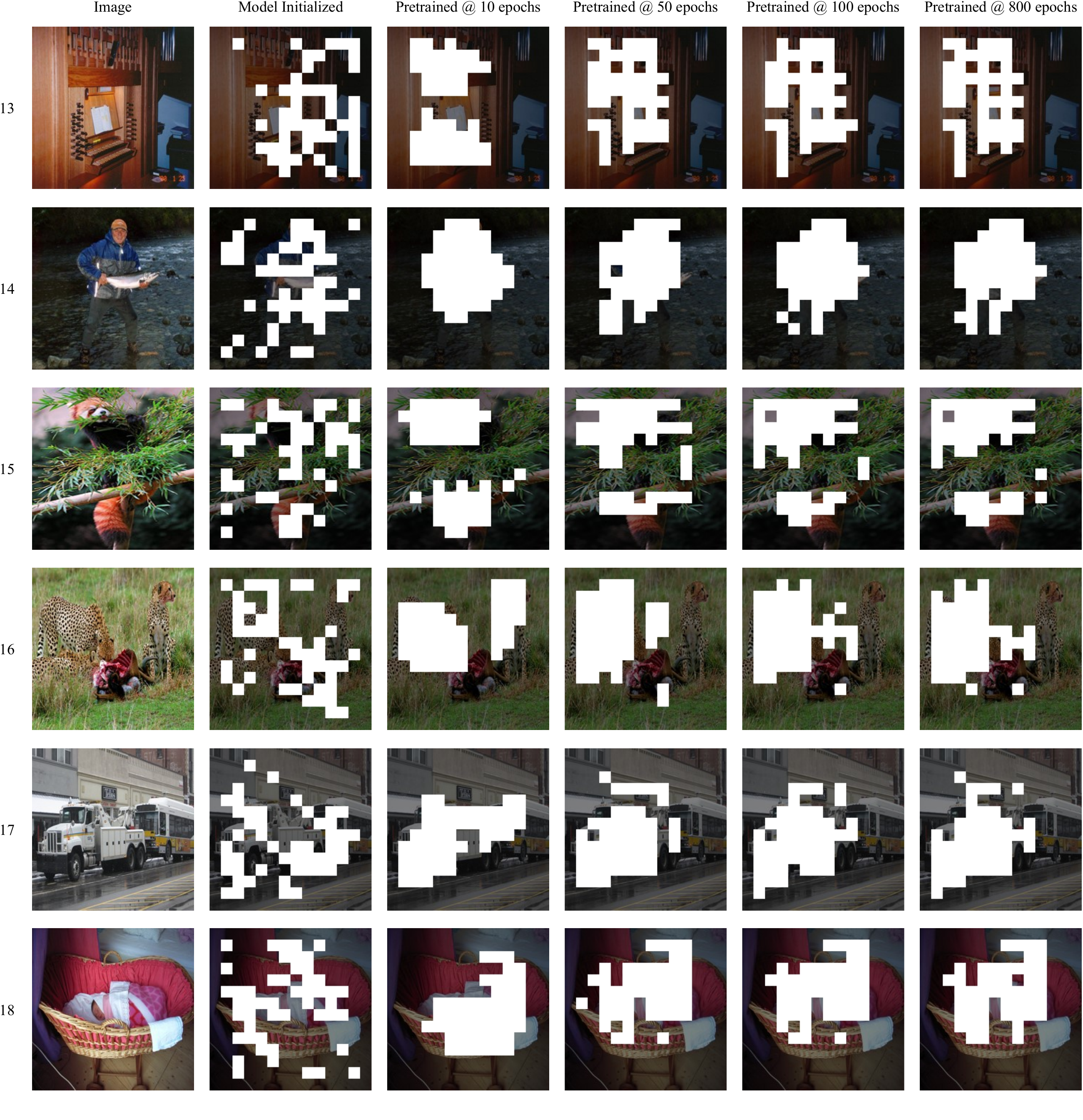}
    \caption{Masked patches by AutoMAE during the pretraining for 800 epochs. White patches are the output of $\log(\mathrm{softmax}(\cdot))$ with the top $25\%$ probabilities.}
    \label{fig:masked3}
    \end{center}
\end{figure*}

After passing the ViT attention feature map to the two-layer CNN generator, the mask is obtained by applying $\log(\mathrm{softmax}(\cdot))$ and Gumbel softmax. Due to the randomness of Gumbel, we take the output of $\log(\mathrm{softmax}(\cdot))$ and highlight those tokens with the top $25\%$ probabilities. They are visualized in Figures~\ref{fig:masked1}-\ref{fig:masked3} as white patches. 

Figures~\ref{fig:masked1}-\ref{fig:masked3} indicate that the mask generator can learn the information of foreground objects. We can see that at the beginning of training, the mask has no specific patterns. After a certain period of training, it becomes focused on the foreground object. It can even follow the shape of discrete parts (see showcases 15 and 16). It shows that the mask tends to fit the shape of the foreground object best at the early stage (see showcases 3, 5-11). For complex scenes in Fig.~\ref{fig:masked3}, it still generates meaningful masks around at least one foreground object. Although the generated mask is gradually changed during the pretraining process, it is mainly focused on the area around the foreground, which indicates that it will have a high probability to be dropped in AutoMAE. It demonstrates that our model can generate semantic masks to improve the self-supervised learning process of the original MAE.

\end{document}